\pdfoutput=1
\documentclass[11pt]{article}

\usepackage{acl}

\usepackage{times}
\usepackage{latexsym}

\usepackage[T1]{fontenc}
\usepackage{kotex}
\usepackage{arabtex}
\usepackage{utf8}
\setcode{utf8}

\usepackage[utf8]{inputenc}

\usepackage{microtype}

\usepackage{comment}

\usepackage{xparse}
\usepackage{graphicx}
\usepackage[boxed]{algorithm2e}

\usepackage{textcomp}
\usepackage{enumitem}
\usepackage{hyperref}
\usepackage{soul}

\NewDocumentCommand{\arafat}
{ mO{} }{\textcolor{red}{\textsuperscript{\textit{Arafat}}\textsf{\textbf{\small[#1]}}}}

\NewDocumentCommand{\yulong}
{ mO{} }{\textcolor{red}{\textsuperscript{\textit{Yulong}}\textsf{\textbf{\small[#1]}}}}
\NewDocumentCommand{\avi}
{ mO{} }{\textcolor{blue}{\textsuperscript{\textit{Avi}}\textsf{\textbf{\small[#1]}}}}

\newcommand{\ie}{\textit{i.e.}}
\newcommand{\eg}{\textit{e.g.}}
\newcommand{\ir}{\textsc{ir}}

\newcommand{\clir}{\textsc{clir}}
\newcommand{\kd}{\textsc{kd}}
\newcommand{\colbert}{\textsc{C}ol\textsc{bert}}
\newcommand{\mt}{\textsc{mt}}
\newcommand{\plm}{\textsc{plm}}
\newcommand{\xlmr}{\textsc{\mbox{xlm-r}}}
\newcommand{\sota}{\textsc{sota}}
\newcommand{\drdecr}{\textsc{Dr.Decr}}{}
\newcommand{\en}{\textsc{en}}

\usepackage{makecell}

\title{Learning Cross-Lingual IR from an English Retriever}

\author{Yulong Li$^{\dag}$\thanks{\quad Equal contribution.}  , Martin Franz$^{\ddag}$$^{*}$, Md Arafat Sultan$^{\ddag}$$^{*}$,\\ \textbf{Bhavani Iyer$^{\ddag}$}, \textbf{Young-Suk Lee$^{\ddag}$} \and \textbf{Avirup Sil$^{\ddag}$}  \\
        $^{\dag}$IBM Research, $^{\ddag}$IBM Research AI \\ \{yulongl, franzm, bsiyer, ysuklee, avi\}@us.ibm.com \\arafat.sultan@ibm.com}

\begin{document}
\maketitle

\begin{abstract}\end{list}
We present \drdecr{}
%(short for, 
(\textbf{D}ense \textbf{R}etrieval with \textbf{D}istillation-\textbf{E}nhanced \textbf{C}ross-Lingual \textbf{R}epresentation), 
%as 
a new cross-lingual information retrieval (\clir{}) system trained using multi-stage knowledge distillation (\kd{}).
The teacher of \drdecr{} relies on a highly effective but computationally expensive two-stage inference process consisting of query translation and monolingual \ir{}, while the student, \drdecr{}, executes a single \clir{} step.
We teach \drdecr{} powerful multilingual representations as well as \clir{} by optimizing two corresponding \kd{} objectives.
Learning useful representations of non-English text from an English-only retriever is accomplished through a cross-lingual token alignment algorithm that relies on the representation capabilities of the underlying multilingual encoders.
In both in-domain and zero-shot out-of-domain evaluation,
%our proposed method 
\drdecr{} demonstrates far superior accuracy over direct fine-tuning with labeled \clir{} data.
%, with gains of 25.4 and 14.9 Recall@5kt, respectively.
%\drdecr{} is also the current\footnote{At the time of writing this paper.} best single-model retriever on the XOR-TyDi leaderboard.
It is also the best single-model retriever on the XOR-TyDi benchmark at the time of this writing.
\end{abstract}

\section{Introduction}
\label{section:introduction}
%Multilingual models are a topic of growing interest in \textsc{NLP} given their criticality to the universal adoption of \textsc{AI}.
%Multilingual models are critical for the adoption of \textsc{AI} on a global scale.
Multilingual models are critical for the democratization of \textsc{AI}.
Cross-lingual information retrieval (\clir) \cite{braschler1999cross,shakery2013leveraging,jiang2020cross, asai-etal-2021-xor}, for example, can find relevant text in a high-resource language such as English even when the query is posed in a different, possibly low-resource, language.
%,shi2021cross,saleh-etal-2019-extended,
In this work, we develop useful \clir{} models for this constrained, yet important, setting where a retrieval corpus is available only in a single high-resource language (English in our experiments).

A straightforward solution to this problem could use machine translation (\mt{}) to translate the query into English, and then perform English \ir{} \cite{asai-etal-2021-xor}.
While such a two-stage process can produce reasonably accurate predictions, an alternative end-to-end approach that can tackle the problem purely cross-lingually, \ie{}, without involving \mt{} for inference, would clearly be more efficient and cost-effective.
Pre-trained multilingual masked language models (\plm{}s) such as multilingual \textsc{bert} \cite{devlin-etal-2019-bert} or {\sc xlm-r}o{\sc bert}a (\xlmr{}) \cite{conneau-etal-2020-unsupervised} can provide the foundation for such a one-step solution, as simply fine-tuning a \plm{} with labeled \clir{} data would yield a cross-lingual retriever \cite{asai-etal-2021-one}.
%\textcolor{red}{Make the argument that this approach has been successful in related problems (with reference).}

Here we first run an empirical evaluation of these two approaches on a public \clir{} benchmark \cite{asai-etal-2021-xor}, which involves both in-domain and zero-shot out-of-domain tests.
We use \mbox{\colbert{}} \cite{khattab-etal-2020-colbert,khattab-etal-2021-relevance}---a state-of-the-art (\sota{}) neural \ir{} model that has been shown to outperform other recent methods such as \textsc{dpr} \cite{karpukhin2020dense}---as our \ir{} architecture and \xlmr{} as the underlying \plm{} for both methods (\textsection{\ref{section:method}}).
%\footnote{Due to its state-of-the-art (\sota{}) performance, outperforming recent models such as DPR \cite{karpukhin2020dense}.}
Results indicate that the \mt{}-based solution can be vastly more effective than direct \en{} + \clir{} fine-tuning, with observed differences of 22.2--28.6 Recall@5k-tokens  (\textsection{\ref{section:experiments}}).
Crucially, the modular design of the former allows it to leverage additional English-only training data for its \ir{} component, providing significant boosts to its performance.

The above findings lead naturally to the central research question of this paper: Can a high-performance \clir{} model be trained that can operate without having to rely on \mt{}?
To answer the question, instead of viewing the \mt{}-based approach as a competing one, we propose to leverage its strength via knowledge distillation (\kd{}) into an end-to-end \clir{} model, which we call \drdecr{} (\textbf{D}ense \textbf{R}etrieval with \textbf{D}istillation-\textbf{E}nhanced \textbf{C}ross-lingual \textbf{R}epresentation).
\kd{} \cite{hinton-etal-2015-distilling} is a powerful supervision technique typically used to distill the knowledge of a large \textit{teacher} model about some task into a smaller \textit{student} model \cite{mukherjee2020xtremedistil,turc2020well}.
Here we propose to use it in a slightly different context, where the teacher and the student retriever are identical in size, but the former has superior performance simply due to utilizing \mt{} output and consequently operating in a high-resource and low-difficulty monolingual environment.
%To the best of our knowledge, \kd{} has not been explored before in this setting in NLP.

\begin{figure*}[h]
\includegraphics[scale=0.5]{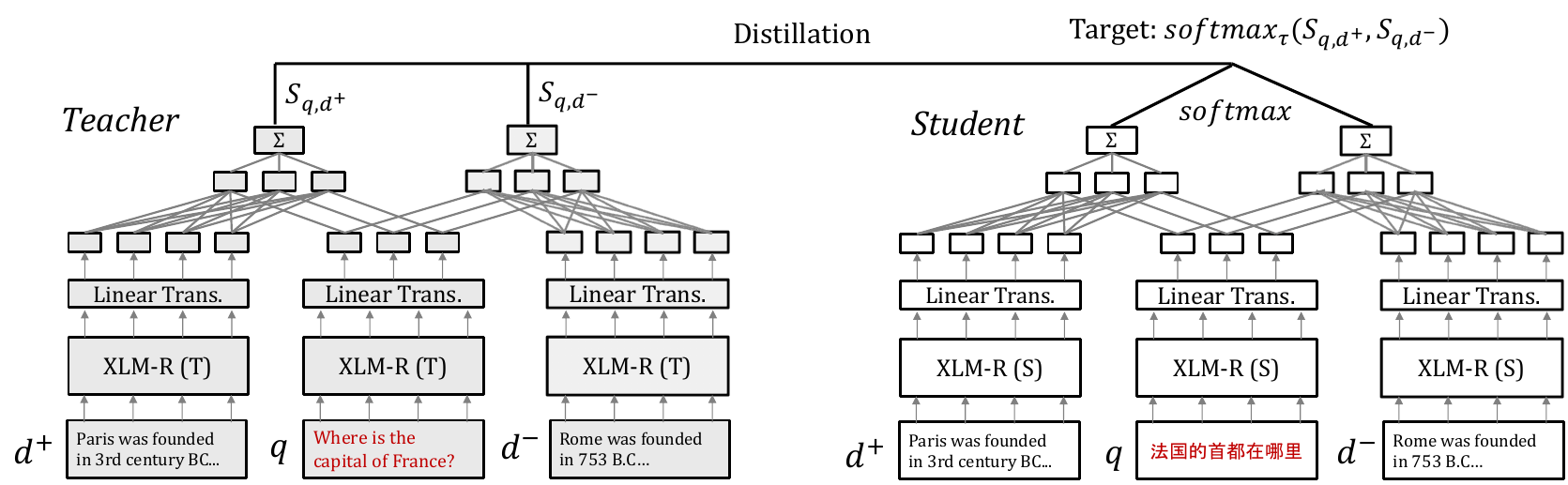}
\centering
\caption{Relevance score distillation. 
The teacher is shown all-English triples while the student's (\drdecr{}) query input is non-English.
Training minimizes the student's KL divergence from the teacher's output softmax distribution over $S_{q,d^+}$ and $S_{q,d^-}$ ($\tau$ is the temperature).
}
\label{figure:relevance-dist}
\end{figure*}

We run two independent \kd{} operations (\textsection{\ref{subsection:KD}}).
One directly optimizes an \ir{} objective by utilizing labeled \clir{} data: parallel questions (English and non-English) and corresponding relevant and non-relevant English passages.
The teacher and the student, \drdecr{}, are shown the English and non-English versions of the questions, respectively; the training objective is for \drdecr{} to match the soft query-passage relevance predictions of the teacher.
The second \kd{} task is representation learning from parallel text, where \drdecr{} learns to encode a non-English text in a way that matches the teacher's encoding of the aligned English text, \textit{at the token level}.
The cross-lingual token alignments needed to create the training data for this task are generated using a greedy alignment process, which exploits the multilingual representation capabilities of the underlying \plm{} encoders.

%Experimental results (\textsection{\ref{section:experiments}}) show that our proposed methods can regain much of the performance loss from the two-stage solution to the end-to-end cross-lingual model. 
%\textcolor{red}{On an XOR-TyDi test set, the student outperforms the cross-lingual \mbox{\colbert{}} baseline by 25.4 points in terms of Recall@5kt, trailing the teacher processing English queries by just 3.2 points.}
In our evaluation on the XOR-TyDi benchmark \cite{asai-etal-2021-xor}, \drdecr{} outperforms the fine-tuned \colbert{} baseline by 25.4 (in-domain) and 14.9 (zero-shot) Recall@5k-tokens, recovering much of the performance loss from the \mt{}-based solution.
It is also the best single-model \ir{} system on the XOR-TyDi leaderboard\footnote{\href{https://nlp.cs.washington.edu/xorqa/}{https://nlp.cs.washington.edu/xorqa/}} at the time of this writing.
Ablation studies show that each of our two \kd{} processes contribute significantly towards the final performance of \drdecr{}.

Our contributions can be summarized as follows: \textbf{(1)} We present an empirical study of the effectiveness of a \sota{} \ir{} method (\mbox{\colbert{}}) on cross-lingual \ir{} with and without \mt{}, \textbf{(2)} We propose a novel end-to-end cross-lingual solution that uses knowledge distillation to learn both improved text representation and retrieval,
\textbf{(3)} We demonstrate with a new cross-lingual alignment algorithm that distillation using parallel text can strongly augment cross-lingual \ir{} training, and %models and can be as effective as <query, positive, negative> triples.
\textbf{(4)} We achieve new single-model \sota{} results on \mbox{XOR-TyDi}.

\begin{comment}
These are the things we hope to be able to show (please add if anything is missing):

\begin{itemize}
    \item Translation + English \ir{} is a more effective approach than direct cross-lingual IR. (Done)
    \item However, it requires two operations at inference time---we want to do a single cross-lingual \ir{} operation and get as close as possible to the two-stage pipelined approach. (Ongoing)
    \item Our basic framework is knowledge distillation (\kd{}); with a simple coarse-grained \kd{} in the final layer, we already do better than the baseline. (Done)
    \item But we introduce additional finer-grained \kd{} mechanisms to further improve results. (Ongoing)
    \item Use of synthetic data also improves performance. (Is this true?)
    \item Optional: Generation of synthetic data for this problem requires a careful approach due to scarcity of training data. -- This I don't think is feasible anymore, because we are using translations so many features of the "careful approach" are actually not being used.
    \item As the boundary between the representations of the sub-word units based on the original query and the "padding" is known, it can be reflected in finer-grained distillation training (\eg{} adjusting the weight, or excluding the "padding" from alignment for the alignment-based approaches).
\end{itemize}
\end{comment}

\section{Method}
\label{section:method}
Here we first describe our base \ir{} architecture (\mbox{\colbert{}}) and then the proposed \kd{}-based cross-lingual training algorithms.

\subsection{The \colbert{} Model}
\label{subsection:colbert}
\mbox{\colbert{}} \cite{khattab-etal-2020-colbert} employs a transformer-based encoder to separately encode the input query and document, followed by a linear compression layer. 
Each training instance is a \mbox{<$q$, $d^+$, $d^-$>} triple, where $q$ is a query, $d^+$ is a positive (relevant) document and $d^-$ is a negative (non-relevant) document. A relevance score $S_{q, d}$ for the pair $(q,d)$ is first computed using Eq.~\ref{equation:colbert}, where $d \in \{d^+, d^-\}$ and $E_{q_i}$ and $E_{d_j}$ are the output embeddings of query token $q_i$ and document token $d_j$, respectively. 
For a given training triple, a cross-entropy loss is minimized for the softmax over $S_{q,d^+}$ and $S_{q,d^-}$. 
%The late interaction between the document and query after the linear layer makes it possible to separate their computations. Therefore 

\begin{equation}
\label{equation:colbert}
%S_q,_d := \sum_{i \in [|E_{q_i}|]} max_{j \in [|E_{d_j}|]}E_{q_i} \cdot E^T_{d_j}
S_q,_d := \sum_{i \in [|q|]} max_{j \in [|d|]}E_{q_i} \cdot E^T_{d_j}
\end{equation}

For inference, the embeddings of all documents are calculated \textit{a priori}, while the query embeddings and the relevance score are computed at runtime.
%This way, the ColBERT model achieved both high effectiveness and low latency. 

\subsection{Knowledge Distillation}
\label{subsection:KD}
Our teacher and \drdecr{} are both \mbox{\colbert{}} models that fine-tune the same underlying multilingual \plm{} for \ir{}.
The teacher is first trained with \mbox{all-English} triples using the procedure of \S\ref{subsection:colbert}.
The goal of the subsequent \kd{} training is to teach \drdecr{} to reproduce the behavior of this teacher when it sees non-English translations of the teacher's English questions.

We apply \kd{} at two different stages of the \mbox{\colbert{}} workflow: (\textit{a}) relevance score computation ($S_{q,d}$ in Eq.~\ref{equation:colbert}), and (\textit{b}) encoding (\eg{}, $E_{q_i}$).
Figure~\ref{figure:relevance-dist} depicts (\textit{a}) in detail,
%, where the teacher is shown an English question and the student its translation in another language; all documents are in English.
where training minimizes the \textsc{KL} divergence between the \drdecr{}'s and the teacher's output softmax distributions (with temperature)  over $S_{q,d^+}$ and $S_{q,d^-}$.

\begin{comment}
\begin{equation}
\label{equation:relevance-distill}
D_{KL}(S \| T ) := \sum_{d \in \{d^+, d^-\}} \frac{e^{-{S_{q, d}}/\tau}}{...} 
\end{equation}
\end{comment}

\begin{comment}
Other than public available dataset, we also generated synthetic triples to improve the training result… \textcolor{red}{to be added}
\end{comment}

\SetKwComment{Comment}{//}{}
\DontPrintSemicolon
\begin{algorithm}[t]
\small
\textbf{Input}: \\
$v_T$: Teacher's representation of tokenized  English (EN) text. \\
$v_S$: Student's representation of parallel non-EN text. \\
\textbf{Output}: \\
$v^{(a)}_T$: Reordered teacher output embeddings to reflect position-wise alignment with $v_S$. \\
\textbf{Procedure}: \\
$DM \gets cosine\_distance(v_T, v_S)$ \Comment{matrix}
\Comment{get index pairs to swap in $v_T$}
$swaps \gets [\ ]$\;
%\For{\_ in range(DM\_size[0])}
\For{row in rows(DM)}{
\Comment{loop runs $|v_T|$ times}
$minValue \gets min(DM)$\;
$i, j \gets index\_of(minValue)$\;
\Comment{swap rows i and j}
$DM[[i,j], :] = DM[[j,i], :]$\;
\Comment{set row $j$ and column $j$ to +$\infty$}
$DM[j, :] \gets + \infty$\;
$DM[:, j] \gets + \infty$\;
$swaps.append((i, j))$
}
\Comment{swap teacher's output tokens}
$v^{(a)}_T \gets v_T$\;
\For{s in swaps}{
$v^{(a)}_T[s[0], s[1]] \gets v_T[s[1], s[0]]$\;
}
\caption{Cross-lingual alignment.\label{algo:cl-align}}
\end{algorithm}

Labeled training data for \clir{} are scarce, whereas \mt{}, being a more established area of research, has produced a large amount of parallel text over the years.
We seek to exploit existing parallel corpora in our second \kd{} training, where we teach \drdecr{} to compute representations of non-English texts that closely match the teacher's representations of aligned English texts.
Importantly, since \mbox{\colbert{}} computes a single vector for each individual input token (\ie{}, a \plm{} vocabulary item) and not for the entire input text, our algorithm must support distillation at the token level.

To achieve this, we design an unsupervised cross-lingual token alignment algorithm.
Assuming $(ne_1, ..., ne_S)$ to be the ordered tuple of tokens in a non-English text and $(e_1, ..., e_T)$ the corresponding tuple from the parallel English text, each iteration of this algorithm greedily picks the next $(ne_i, e_j)$ pair with the highest cosine similarity of their output embeddings.
Algorithm~\ref{algo:cl-align} implements this idea by repositioning the teacher's tokens so that they are position-wise aligned with the corresponding \drdecr{} tokens.
Note that the design choice of fine-tuning a common multilingual \plm{} for the teacher and the \drdecr{}, even though the former is tasked with only handling English content, is key for this algorithm as it relies on the \plm{}s' multilingual representation capabilities. See Appendix \ref{appendix:prepreocessing} for details on our parallel corpora used for training.

\begin{figure}[t]
\includegraphics[scale=0.4]{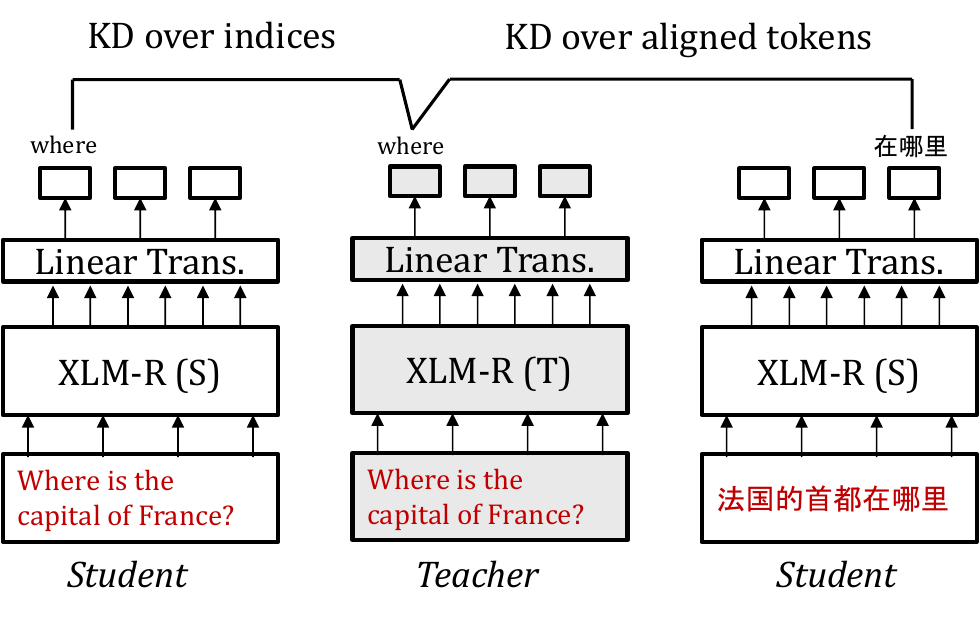}
\centering
\caption{Distillation for representation learning. 
The student (\drdecr{}) learns to encode both English and non-English tokens in context that matches the teacher's output embeddings for corresponding English tokens.}
\label{figure:output-dist}
\end{figure}

In addition to cross-lingual alignment, we also perform a similar \kd{} procedure in which both the teacher and the \drdecr{} are shown the same English text.
This step is useful because \mbox{\colbert{}} uses a shared encoder for the query and the document, necessitating a student that is able to effectively encode text from both English documents and non-English queries.

Using the alignment information, we train \drdecr{} by minimizing the Euclidean distance between its representation of a token (English or non-English) and the teacher's representation of the corresponding English token.
Figure~\ref{figure:output-dist} shows the \kd{} process for representation learning.

\begin{comment}
\begin{itemize}
    \item Text describing: overall approach, level 3 \kd{}, level 1 \kd{}, synthetic data generation
    \item Diagram of approach
    \item Algorithms: main, cross-lingual alignment
\end{itemize}
\end{comment}

\section{Experiments}
\label{section:experiments}

\begin{table*}[t]
\small
\centering
%\begin{tabular}{lcc}
\begin{tabular}{l|cc}
%\multicolumn{1}{c}{\textbf{System}} & \textbf{Dev} & \textbf{Test} \\
\multicolumn{1}{c}{\textbf{System}} & \textbf{R@5kt} & \textbf{R@2kt} \\
\hline
%\Xhline{2\arrayrulewidth}
\multicolumn{1}{l}{\textit{With target domain supervision:}} \\
\hline
%ColBERT: $\textnormal{FT}_{XOR}$ & ? \\
%Baseline 2: NQ FT $\rightarrow$ XOR FT & 45.6 & 47.7 \\
~~$\textnormal{\colbert{}}_{\textsc{cl}}$: \textit{ft}({\textsc{xor}}) & 32.9 & 23.9 \\
~~$\textnormal{\colbert{}}_{\textsc{en+cl}}$: \textit{ft}({\textsc{nq}}) $\rightarrow$ \textit{ft}({\textsc{xor}}) & 47.7 & 38.1 \\
%Teacher: NQ FT + EN Questions & 73.7 & 76.3 \\
~~Teacher: \textsc{MT} $+$ $\textnormal{\colbert{}}_{\textsc{en}}$ & 76.3 & 70.5 \\
%Baseline -> XOR & 59.3 & 0.0 \\
%Baseline -> Synthetic data -> XOR & 61.4 & 66.1 \\
%Baseline -> Parallel Corpus \\
%-> Synthetic data -> XOR & 67.7 & 72.6 \\
%Baseline $\rightarrow$ KD (PC) $\rightarrow$ KD (XOR) & 67.6 & 73.1 \\
~~\drdecr{}: $\textnormal{\colbert{}}_{\textsc{en+cl}}$ $\rightarrow$ $\textsc{KD}_{\textsc{pc}}$ $\rightarrow$ $\textsc{KD}_{\textsc{xor}}$ & 73.1 & 66.0 \\
\hline
\multicolumn{1}{l}{\textit{Zero-shot:}} \\
\hline
~~$\textnormal{\colbert{}}_{\textsc{cl}}$: \textit{ft}({\textsc{mkqa}}) & 23.6 & 16.7 \\
~~$\textnormal{\colbert{}}_{\textsc{en+cl}}$: \textit{ft}({\textsc{nq}}) $\rightarrow$ \textit{ft}({\textsc{mkqa}}) & 46.9 & 38.7 \\
~~Teacher: \textsc{MT} $+$ $\textnormal{\colbert{}}_{\textsc{en}}$ & 69.1 & 62.7 \\
~~\drdecr{}: $\textnormal{\colbert{}}_{\textsc{en+cl}}$ $\rightarrow$ $\textsc{KD}_{\textsc{pc}}$ $\rightarrow$ $\textsc{KD}_{\textsc{mkqa}}$ & 61.8 & 54.3 \\
%\Xhline{2\arrayrulewidth}
\hline
\end{tabular}
\caption{
Performance on the XOR-TyDi test set.
\textit{ft}: fine-tuning; \textit{CL}: cross-lingual; \textsc{nq}: the Natural Questions train set; \textsc{pc}: parallel corpus; \textsc{xor}: the XOR-TyDi train set.
%; \kd{}: knowledge distillation.
Direct fine-tuning of \colbert{} with \ir{} triples underperforms \mbox{\mt{} $+$ English \ir{}} by 22.2--32.4 points; the proposed \kd{}-based methods close this gap by 65.0$\%$--88.8$\%$.
%All ColBERT models fine-tune an XLM-RoBERTa base masked language model.
}
\label{table:main-results}
\end{table*}

%\textcolor{darkblue}{\textbf{\textit{Setup.}}}
\subsection{Setup}
Our primary \clir{} dataset is XOR-TyDi \cite{asai-etal-2021-xor}, which contains examples in seven typologically diverse languages: Arabic (Ar), Bengali (Bn), Finnish (Fi), Japanese (Ja),
Korean (Ko), Russian (Ru) and Telugu (Te).
For standard in-domain experiments, we use a train-dev-test split of this dataset.
There are 2,113 questions in the test set.
For zero-shot experiments, we use the MKQA \cite{mkqa} dataset for training and validation, and the following shared languages in the XOR-TyDi test set for evaluation: Ar, Fi, Ja, Ko and Ru.
Both training sets contain English questions and their human translations in the other languages, their short answers and corresponding relevant (positive) and non-relevant (negative) Wikipedia snippets.
Additionally, we use training examples from the Natural Questions (NQ) dataset ~\cite{kwiatkowski-etal-2019-nq} for English pre-training of the baseline model.
Further details on data pre-processing and the final training sets 
%and the optimal hyperparameter configurations 
are provided in Appendix \ref{appendix:prepreocessing}.
%Appendices \ref{appendix:prepreocessing} and \ref{appendix:hyperparameters}.

The \clir{} baseline used in our experiments is \colbert{} with an underlying \xlmr{} \plm{}, which we iteratively fine-tune first on English and then on cross-lingual \ir{} triples for optimal performance.
Our \drdecr{} model is initialized with the parameter weights of this baseline, and is further fine-tuned using the two \kd{} objectives.
The \kd{} teacher is a \colbert{} model fine-tuned with only English triples, as stated before.
During evaluation, it is given machine-translated questions that come with the XOR-TyDi dataset.
Appendices \ref{appendix:prepreocessing} and  \ref{appendix:hyperparameters} contain additional details on the supervision of these models and the optimal hyperparameter configurations.

We evaluate using Recall at $t$ tokens for $t \in \{2000, 5000\}$, \ie{}, R@2kt and R@5kt \cite{asai-etal-2021-xor}, which compute the fraction of questions for which the ground truth short answer is contained within the top $t$ tokens of the retrieved passages.
%Appendix~\ref{appendix:r@2kt} also presents R@2kt results.

\begin{table}[t]
\centering
\small
\begin{tabular}{lcc}
%\begin{tabular}{l|c}
%\multicolumn{1}{c}{\textbf{System}} & \textbf{Dev} & \textbf{Test} \\
\multicolumn{1}{c}{\textbf{Language}} & \textbf{Baseline} & \textbf{\drdecr{}} \\
%\Xhline{2\arrayrulewidth}
\hline
\multicolumn{3}{l}{\textit{With target domain supervision:}} \\
\hline
Te & 63.0 & 83.2 \\
Bn & 53.3 & 85.9 \\
Fi & 49.4 & 69.4 \\
Ja & 39.4 & 65.1 \\
Ko & 44.9 & 68.8 \\
Ru & 39.2 & 68.8 \\
Ar & 44.3 & 70.2 \\
\hline
\textbf{Avg} & \textbf{47.7} & \textbf{73.1} \\
%\hline
%\\
\hline
\multicolumn{3}{l}{\textit{Zero-shot:}} \\
\hline
Fi & 55.4 & 66.9 \\
Ja & 44.0 & 58.5 \\
Ko & 48.4 & 62.8 \\
Ru & 41.4 & 57.8 \\
Ar & 45.3 & 61.8 \\
\hline
\textbf{Avg} & \textbf{46.9} & \textbf{61.8} \\
%\hline
\end{tabular}
\caption{
R@5kt scores for in-domain and zero-shot evaluation on individual languages. Baseline for the target domain experiment: ~~$\textnormal{\colbert{}}_{\textsc{en+cl}}$: \textit{ft}({\textsc{nq}}) $\rightarrow$ \textit{ft}({\textsc{xor}}). Baseline for the zero-shot experiment: ~~$\textnormal{\colbert{}}_{\textsc{en+cl}}$: \textit{ft}({\textsc{nq}}) $\rightarrow$ \textit{ft}({\textsc{mkqa}})
}
\label{table:individual-lang-results}
\end{table}

\begin{table*}[h]
\centering
\small
%\begin{tabular}{lcc}
\begin{tabular}{l|cc}
%\multicolumn{1}{c}{\textbf{System}} & \textbf{Dev} & \textbf{Test} \\
\multicolumn{1}{c}{\textbf{System}} & \textbf{R@5kt} & \textbf{R@2kt} \\
%\Xhline{2\arrayrulewidth}
\hline
\multicolumn{1}{l}{\textit{With target domain supervision:}} \\
\hline
~~\drdecr{}: $\textnormal{\colbert{}}_{\textsc{en+cl}}$ $\rightarrow$ $\textsc{KD}_{\textsc{pc}}$ $\rightarrow$ $\textsc{KD}_{\textsc{xor}}$ & 73.1 & 66.0 \\
%Baseline 2 $\rightarrow$ KD (XOR) & 59.3 & 63.6 \\
~~$\textnormal{\colbert{}}_{\textsc{en+cl}}$ $\rightarrow$ $\textsc{KD}_{\textsc{pc}}$ & 68.6 & 60.6 \\
~~$\textnormal{\colbert{}}_{\textsc{en+cl}}$ $\rightarrow$ $\textsc{KD}_{\textsc{xor}}$ & 63.6 & 56.6 \\
%Baseline 2 $\rightarrow$ KD (PC) & 63.9 & 68.6 \\
~~$\textnormal{\colbert{}}_{\textsc{en+cl}}$ & 47.7 & 38.1 \\
%Baseline -> Parallel corpus -> XOR & 67.6 & 73.1 \\s
%\Xhline{2\arrayrulewidth}
\hline
\multicolumn{1}{l}{\textit{Zero-shot:}} \\
\hline
~~\drdecr{}: $\textnormal{\colbert{}}_{\textsc{en+cl}}$ $\rightarrow$ $\textsc{KD}_{\textsc{pc}}$ $\rightarrow$ $\textsc{KD}_{\textsc{mkqa}}$ & 61.8 & 54.3 \\
~~$\textnormal{\colbert{}}_{\textsc{en+cl}}$ $\rightarrow$ $\textsc{KD}_{\textsc{pc}}$ & 55.9 & 47.7 \\
~~$\textnormal{\colbert{}}_{\textsc{en+cl}}$ $\rightarrow$ $\textsc{KD}_{\textsc{mkqa}}$ & 49.3 & 40.9 \\
%Baseline 2 $\rightarrow$ KD (PC) & 63.9 & 68.6 \\
~~$\textnormal{\colbert{}}_{\textsc{en+cl}}$ & 46.9 & 38.7 \\
\hline
\end{tabular}
\caption{
Results of the ablation study. 
\kd{} with parallel corpus ($\textsc{KD}_{\textsc{pc}}$) and \ir{} triples ($\textsc{KD}_{\textsc{xor}}$) both play key roles in our \drdecr{} model.
Interestingly, the former has a greater impact on the model's performance. 
}
\label{table:ablation-results}
\end{table*}

\subsection{Evaluation}
%\noindent\textbf{\textit{Evaluation.}}
%\textcolor{darkblue}{\textbf{\textit{Evaluation.}}}
Table~\ref{table:main-results} compares the performance of our different models.
%measured by their R@5kt scores.
First, looking at the R@5kt results, we observe that pre-training the baseline model with English \ir{} triples from the NQ train set (rows 2, 6)
%~\cite{kwiatkowski-etal-2019-nq}
substantially boosts its performance in both in-domain and zero-shot settings.
However, it still underperforms the \mt{} $+$ English \ir{} pipeline (rows 3, 7) by 28.6 and 22.2 points, respectively. 
By distilling first with the parallel corpus (for representation learning) and then with the \ir{} triples (for \clir{}), \drdecr{} (row 4) yields an improvement of 25.4 points over the baseline model in in-domain evaluation, which, quite impressively, is within 3.2 points of the teacher's score.
A sizable gain of 14.9 points is also observed in zero-shot evaluation (row 8).
Finally, the R@2kt numbers show a very similar pattern.
%We present more fine-grained per-language results in Appendix~\ref{appendix:language-results}.%\newline

Table~\ref{table:individual-lang-results} shows the performance (R@5kt) of \drdecr{} and the baseline on each individual language:
the former outperforms the latter both with and without target domain supervision, yielding large gains across all languages.
These results demonstrate the robustness of our approach, which stems from combining the individual strengths of \mt{}, English \ir{} and \kd{} in a single model.

\subsection{Leaderboard Submission}
%\textcolor{darkblue}{\textbf{\textit{Leaderboard Submission.}}}
The \drdecr{} model trained on the XOR-TyDi training set, shown in Table~\ref{table:main-results} row 4, is the best single-model retriever on the XOR-TyDi leaderboard\footnote{\href{https://nlp.cs.washington.edu/xorqa/}{https://nlp.cs.washington.edu/xorqa/}} at the time of this writing. Since our parallel corpus extraction process relies on in-house source code that is not publicly available, we submitted to the ``Systems using External APIs'' category.
Crucially, all other submitted systems under the External APIs category rely on \mt{} at decoding time, avoiding which is one of the primary goals of our work. We also created parallel corpora purely from public available sources.\footnote{\href{https://opus.nlpl.eu}{https://opus.nlpl.eu}} Our model distilled with these instances also achieved top position on the white-box systems leaderboard of XOR-TyDi.
%Please see Appendix \ref{appendix:leaderboard} for further details.

%\subsection{Ablation Study}
%\noindent\textbf{\textit{Ablation Study.}}
%To study the effect of the two different \kd{} operations on our student model's performance, we train two additional students. 
%Each of these students goes through only one of the two \kd{} training steps.

\subsection{Ablation Study}
%\textcolor{darkblue}{\textbf{\textit{Ablation Study.}}}
We experiment with two more student models, one distilled with only \clir{} examples and the other with only the parallel corpus.
%Table 2 summarizes the results: \kd{} with only \clir{} examples and with only the parallel corpus improves the system's score by 15.9 and 20.9 points, respectively with target domain supervision.
As the results in Table~\ref{table:ablation-results} show, each has a substantial impact on system performance.
Interestingly, although the parallel corpus does not provide any \ir{} signal, it contributes more to the model's accuracy.
These results also confirm that our cross-lingual alignment algorithm does indeed produce useful alignments.

\begin{comment}
 \begin{itemize}
 \item Datasets.
 \item Baselines.
 \item Results on XOR.
 \item Zero-shot evaluation on MKQA(?)
 \item Ablation results
 \item Maybe some qualitative analysis
 \end{itemize}

Table~\ref{table:main-results}...
\end{comment}

\begin{comment}
\begin{table}
\centering
\begin{tabular}{lc}
\multicolumn{1}{c}{\textbf{System}} & \textbf{Test} \\
\hline
\hline
BM25 + T(Q)? & 0.0 \\
ColBERT CL & 0.0 \\
\hline
ColBERT Eng + T(Q) & 0.0 \\
ColBERT KD & 0.0 \\
\hline
\hline
\end{tabular}
\caption{
Zero-shot results on MKQA(?).
}
\label{table:zero-shot-results}
\end{table}
\end{comment}

\section{Conclusion}
\label{section:conclusion}
%We show that without the help of machine translation (\mt{}) at inference time, a state-of-the-art IR framework falters on cross-lingual \ir{}.  
%As a solution, we propose distilling the knowledge of a teacher model that performs monolingual \ir{} on \mt{} output into a cross-lingual student model capable of operating without \mt{}.
We train highly effective end-to-end cross-lingual \ir{} models by distilling the knowledge of an English retriever.
We propose separate processes to teach \ir{} and multilingual text representations, and present for the latter a cross-lingual alignment algorithm that only relies on the underlying masked language model's multilingual representation capabilities.
Supervised and zero-shot evaluations show that our model recovers much of the performance lost due to operating in an efficient cross-lingual mode.
Our \kd{}-based method also yields new single-model \sota{} results on the XOR-TyDi benchmark.
Future work will explore \ir{} on unseen languages and evaluation on additional datasets.

\section{Ethics}
\label{section:Ethics}
\subsection{Limitations}
We show the effectiveness of multi-stage knowledge distillation and cross-lingual token alignment in training a cross-lingual information retrieval system. We believe that it can be transferred to more datasets and languages, but here we only show proof of concept for the XOR-TyDi and MKQA datasets and the seven languages mentioned in the paper. 

\subsection{Risks}
The intent of this work is to develop a new method for high-performance cross-lingual information retrieval. It is possible that a malicious user could try to attack the system by providing poor or offensive training data. We do not support it being used in such a manner. The risks of our system are the same as other \textsc{nlp} systems and we do not believe we introduce any additional risk.

\section*{Acknowledgements}
We thank Graeme Blackwood and Christoph Tillmann for providing the in-house parallel corpora. We also thank Akari Asai for her help submitting \drdecr{} to the XOR-TyDi leaderboard.

\begin{comment}
\section{Lit Review}
\begin{itemize}
\item Asai et al. (Title): Created the XOR dataset.
\item Zhuolin Jiang et al. Cross-lingual Information Retrieval with BERT
\end{itemize}
\end{comment}

\begin{comment}
\section*{Acknowledgements}
This document has been adapted
by Steven Bethard, Ryan Cotterell and Rui Yan
from the instructions for earlier ACL and NAACL proceedings, including those for 
ACL 2019 by Douwe Kiela and Ivan Vuli\'{c},
NAACL 2019 by Stephanie Lukin and Alla Roskovskaya, 
ACL 2018 by Shay Cohen, Kevin Gimpel, and Wei Lu, 
NAACL 2018 by Margaret Mitchell and Stephanie Lukin,
Bib\TeX{} suggestions for (NA)ACL 2017/2018 from Jason Eisner,
ACL 2017 by Dan Gildea and Min-Yen Kan, 
NAACL 2017 by Margaret Mitchell, 
ACL 2012 by Maggie Li and Michael White, 
ACL 2010 by Jing-Shin Chang and Philipp Koehn, 
ACL 2008 by Johanna D. Moore, Simone Teufel, James Allan, and Sadaoki Furui, 
ACL 2005 by Hwee Tou Ng and Kemal Oflazer, 
ACL 2002 by Eugene Charniak and Dekang Lin, 
and earlier ACL and EACL formats written by several people, including
John Chen, Henry S. Thompson and Donald Walker.
Additional elements were taken from the formatting instructions of the \emph{International Joint Conference on Artificial Intelligence} and the \emph{Conference on Computer Vision and Pattern Recognition}.
\end{comment}

\bibliography{anthology,custom}

\begin{thebibliography}{15}
\expandafter\ifx\csname natexlab\endcsname\relax\def\natexlab#1{#1}\fi

\bibitem[{Asai et~al.(2021{\natexlab{a}})Asai, Kasai, Clark, Lee, Choi, and
  Hajishirzi}]{asai-etal-2021-xor}
Akari Asai, Jungo Kasai, Jonathan Clark, Kenton Lee, Eunsol Choi, and Hannaneh
  Hajishirzi. 2021{\natexlab{a}}.
\newblock {XOR QA: Cross-lingual Open-Retrieval Question Answering}.
\newblock In \emph{NAACL}.

\bibitem[{Asai et~al.(2021{\natexlab{b}})Asai, Yu, Kasai, and
  Hajishirzi}]{asai-etal-2021-one}
Akari Asai, Xinyan Yu, Jungo Kasai, and Hannaneh Hajishirzi.
  2021{\natexlab{b}}.
\newblock {One Question Answering Model for Many Languages with Cross-lingual
  Dense Passage Retrieval}.
\newblock In \emph{NeurIPS}.

\bibitem[{Braschler et~al.(1999)Braschler, Krause, Peters, and
  Sch{\"a}uble}]{braschler1999cross}
Martin Braschler, J{\"u}rgen Krause, Carol Peters, and Peter Sch{\"a}uble.
  1999.
\newblock {Cross-language Information Retrieval (CLIR) Track Overview}.
\newblock In \emph{TREC}.

\bibitem[{Conneau et~al.(2020)Conneau, Khandelwal, Goyal, Chaudhary, Wenzek,
  Guzm{\'a}n, Grave, Ott, Zettlemoyer, and
  Stoyanov}]{conneau-etal-2020-unsupervised}
Alexis Conneau, Kartikay Khandelwal, Naman Goyal, Vishrav Chaudhary, Guillaume
  Wenzek, Francisco Guzm{\'a}n, Edouard Grave, Myle Ott, Luke Zettlemoyer, and
  Veselin Stoyanov. 2020.
\newblock {{Unsupervised Cross-lingual Representation Learning at Scale}}.
\newblock In \emph{ACL}.

\bibitem[{Devlin et~al.(2019)Devlin, Chang, Lee, and
  Toutanova}]{devlin-etal-2019-bert}
Jacob Devlin, Ming-Wei Chang, Kenton Lee, and Kristina Toutanova. 2019.
\newblock {{BERT: Pre-training of Deep Bidirectional Transformers for Language
  Understanding}}.
\newblock In \emph{NAACL}.

\bibitem[{Hinton et~al.(2014)Hinton, Vinyals, and
  Dean}]{hinton-etal-2015-distilling}
Geoffrey Hinton, Oriol Vinyals, and Jeff Dean. 2014.
\newblock {Distilling the Knowledge in a Neural Network}.
\newblock In \emph{NeurIPS Deep Learning Workshop}.

\bibitem[{Jiang et~al.(2020)Jiang, El-Jaroudi, Hartmann, Karakos, and
  Zhao}]{jiang2020cross}
Zhuolin Jiang, Amro El-Jaroudi, William Hartmann, Damianos Karakos, and Lingjun
  Zhao. 2020.
\newblock {Cross-lingual Information Retrieval with BERT}.
\newblock \emph{arXiv preprint arXiv:2004.13005}.

\bibitem[{Karpukhin et~al.(2020)Karpukhin, Oguz, Min, Lewis, Wu, Edunov, Chen,
  and Yih}]{karpukhin2020dense}
Vladimir Karpukhin, Barlas Oguz, Sewon Min, Patrick Lewis, Ledell Wu, Sergey
  Edunov, Danqi Chen, and Wen-tau Yih. 2020.
\newblock {Dense Passage Retrieval for Open-Domain Question Answering}.
\newblock In \emph{Proceedings of the 2020 Conference on Empirical Methods in
  Natural Language Processing (EMNLP)}.

\bibitem[{Khattab et~al.(2021)Khattab, Potts, and
  Zaharia}]{khattab-etal-2021-relevance}
Omar Khattab, Christopher Potts, and Matei Zaharia. 2021.
\newblock Relevance-guided supervision for openqa with colbert.
\newblock \emph{Transactions of the ACL}.

\bibitem[{Khattab and Zaharia(2020)}]{khattab-etal-2020-colbert}
Omar Khattab and Matei Zaharia. 2020.
\newblock Colbert: Efficient and effective passage search via contextualized
  late interaction over bert.
\newblock In \emph{SIGIR}.

\bibitem[{Kwiatkowski et~al.(2019)Kwiatkowski, Palomaki, Redfield, Collins,
  Parikh, Alberti, Epstein, Polosukhin, Kelcey, Devlin, Lee, Toutanova, Jones,
  Chang, Dai, Uszkoreit, Le, and Petrov}]{kwiatkowski-etal-2019-nq}
Tom Kwiatkowski, Jennimaria Palomaki, Olivia Redfield, Michael Collins, Ankur
  Parikh, Chris Alberti, Danielle Epstein, Illia Polosukhin, Matthew Kelcey,
  Jacob Devlin, Kenton Lee, Kristina~N. Toutanova, Llion Jones, Ming-Wei Chang,
  Andrew Dai, Jakob Uszkoreit, Quoc Le, and Slav Petrov. 2019.
\newblock Natural questions: a benchmark for question answering research.
\newblock \emph{Transactions of the Association of Computational Linguistics}.

\bibitem[{Longpre et~al.(2020)Longpre, Lu, and Daiber}]{mkqa}
Shayne Longpre, Yi~Lu, and Joachim Daiber. 2020.
\newblock {MKQA: A Linguistically Diverse Benchmark for Multilingual Open
  Domain Question Answering}.
\newblock \emph{arXiv preprint arXiv:2007.15207}.

\bibitem[{Mukherjee and Awadallah(2020)}]{mukherjee2020xtremedistil}
Subhabrata Mukherjee and Ahmed Awadallah. 2020.
\newblock {XtremeDistil: Multi-stage Distillation for Massive Multilingual
  Models}.
\newblock In \emph{ACL}.

\bibitem[{Shakery and Zhai(2013)}]{shakery2013leveraging}
Azadeh Shakery and ChengXiang Zhai. 2013.
\newblock {Leveraging Comparable Corpora for Cross-Lingual Information
  Retrieval in Resource-lean Language Pairs}.
\newblock \emph{Information retrieval}, 16(1):1--29.

\bibitem[{Turc et~al.(2020)Turc, Chang, Lee, and Toutanova}]{turc2020well}
Iulia Turc, Ming-Wei Chang, Kenton Lee, and Kristina Toutanova. 2020.
\newblock {Well-Read Students Learn Better: On the Importance of Pre-training
  Compact Models}.
\newblock In \emph{ICLR}.

\end{thebibliography}
\bibliographystyle{acl_natbib}

\newpage
\clearpage
\appendix

\section{Appendix}
\label{sec:appendix}

\subsection{Data Pre-processing}
\label{appendix:prepreocessing}

The official XOR-TyDi training set consists of 15,221 natural language queries, their short answers, and examples of corresponding relevant (positive) and non-relevant (negative) Wikipedia snippets.
%with respect to the queries.  
For most queries, there are one positive and three negative examples.
We remove the 1,699 (11\%) questions that have no answers in the dataset.
%In our experiments, 
A random selection of 90\% of the remaining examples is used for training and the rest for validation.

Following the original XOR-TyDi process, we also obtain additional training examples by running BM25-based retrieval against a Wikipedia corpus and using answer string match as the relevance criterion.
These examples are added to the original set to obtain three positive and 100 negative examples per query.
%, each of the positive examples is used with 100 negative examples in training triples.
%We use pseudo-randomly selected 90\% of the XOR TyDi Leaderboard training set questions for training, the remaining 10\% for hyper-parameter tuning.
As the blind test set for final evaluation, we use the 2,113 questions in the official XOR-TyDi dev set.

Our monolingual (English) training data containing about 17.5\textsc{m} triples are derived from the third fine-tuning round (\colbert{}-QA3) of \colbert{} relevance-guided supervision \cite{khattab-etal-2021-relevance} with NQ examples \cite{kwiatkowski-etal-2019-nq}. %There are about 17.5M (query, positive, negative) triples in this set.
%We use an in-house IBM neural MT system for question translation.

The parallel corpus used in our \kd{} experiments (Table~\ref{table:main-results}) for representation learning is constructed from three different sources: (1) an in-house crawl of Korean, (2) LDC releases (Arabic), and (3) OPUS.\footnote{\href{https://opus.nlpl.eu}{https://opus.nlpl.eu}}
The corpus has a total of 6.9M passage pairs which include .9M pairs in Telugu and 1M pairs in each of the other six languages. The parallel corpus used in our white-box system  was created purely from OPUS. The statistics and sources are shown in the table below.

\begin{table}[h]
\small
\centering
\begin{tabular}{ccc}
\multicolumn{1}{c} {\textbf{Language}} & \textbf{Amount (M)} & \textbf{Source} \\
\hline
Ja & 0.9 & WikiMatrix \\
Ru & 1.7 & WikiMatrix \\
Ar & 1.0 & WikiMatrix \\
Te & 0.7 & WikiMatrix + CCAligned \\
Bn & 1.3 & WikiMatrix + CCMatrix \\
Fi & 1.4 & WikiMatrix + CCMatrix \\
Ko & 1.3 & WikiMatrix + CCMatrix \\
\hline
\end{tabular}
\caption{
Statistic of parallel corpus used in the XOR-TyDi white-box system.
}
\end{table}

For zero-shot experiments,  the training examples are derived from MKQA \cite{mkqa}, which consists of 10k queries selected from NQ, human translated into 25 additional languages, five of which overlap with XOR-TyDI: Ar, Fi, Ja, Ko and Ru.  We construct training data (triples) from 2,037 queries translated into these five languages for which there are corresponding positive and negative passages in the NQ dataset. For each of the five languages, there are 519k triples for a total of 2.6M triples. We set aside 200 queries translated into the 5 languages for a total of 1,000 queries as a development set.  We remove all MKQA queries from the NQ training data for these experiments.
%We use an \texttt{anonymized} neural \mt{} system for question translation that the teacher model relies on .

The \clir{} baseline for our experiments is a \colbert{} model with an \xlmr{} \plm{}, which we first fine-tune with 17.5M NQ examples for one epoch and then 2.9M XOR-TyDi triples for five epochs.
Our \drdecr{} model is initialized with the parameter weights of the baseline, and is further fine-tuned using the two \kd{} objectives. 
The monolingual teacher model---also a \colbert{} model running on top of the pre-trained \xlmr{}---is trained with only the 17.5M NQ triples for one epoch.

\subsection{Model Selection}
\label{appendix:hyperparameters}
All the models were trained with single Nvidia A100 GPU. The longest training time for a single model was less than 200 hours. Following are the final hyperparameter configurations of our different models.
They were selected based on the respective validation sets performance.

\begin{table}[h]
\small
\centering
\begin{tabular}{l|c}
\multicolumn{1}{c}{\textbf{Hyperparameter}} & \textbf{Value} \\
\hline
%\hline
\multicolumn{2}{l}{Standard ColBERT hyperparameters:} \\
\hline
batch size & 192 \\
gradient accumulation steps & 6 \\
linear compression dim & 128 \\
query maxlen & 32 \\
document maxlen & 180 \\
\hline
\multicolumn{2}{c}{} \\
\multicolumn{2}{l}{\textbf{\textit{Target domain supervision}}} \\
\hline
\multicolumn{2}{l}{Baseline model:} \\
\hline
lr (NQ) & 1.5e-6 \\
lr (XOR) & 6e-6 \\
$\#$ Epochs (NQ) & 1 \\
$\#$ Epochs (XOR) & 5 \\
\hline
\multicolumn{2}{l}{Knowledge distillation:} \\
\hline
loss function (XOR) & KLDiv \\
loss function (Parallel corpus) & MSE \\ 
\kd{} temperature (XOR) & 2 \\
% temperature(Synthetic data) & 4 \\
lr (XOR) & 6e-6 \\
%lr (Synthetic data) & 1.5e-6 \\
lr (Parallel corpus) & 4.8e-5 \\
$\#$ Epochs (XOR) & 5 \\
%$\#$ Epochs (Synthetic data) & 1 \\
$\#$ Epochs (Parallel corpus) & 2 \\
\hline
\multicolumn{2}{c}{} \\
\multicolumn{2}{l}{\textbf{\textit{Zero-shot}}} \\
\hline
\multicolumn{2}{l}{Baseline model:} \\
\hline
lr (NQ) & 1.5e-6 \\
lr (MKQA) & 6e-6 \\
$\#$ Epochs (NQ) & 1 \\
$\#$ Epochs (MKQA) & 1 \\
\hline
\multicolumn{2}{l}{Knowledge distillation:} \\
\hline
loss function (MKQA) & KLDiv \\
loss function (Parallel corpus) & MSE \\ 
\kd{} temperature (MKQA) & 1 \\
% temperature(Synthetic data) & 4 \\
lr (MKQA) & 3e-7 \\
%lr (Synthetic data) & 1.5e-6 \\
lr (Parallel corpus) & 2.4e-5 \\
$\#$ Epochs (MKQA) & 1 \\
%$\#$ Epochs (Synthetic data) & 1 \\
$\#$ Epochs (Parallel corpus) & 2 \\
\hline
%\hline
\end{tabular}
\caption{
Hyperparameters used in our test set runs.
}
\label{table:zero-shot-results}
\end{table}

\subsection{Qualitative Analysis}
\label{appendix:qualitative}
To find out what exact weaknesses of the baseline model the proposed method helps to address, we examine thirty random zero-shot test examples where the baseline fails to retrieve the correct answer in the top 5k tokens, but \drdecr{} succeeds within the top 3 passages.
We show four examples in Table~\ref{table:qualitative} with human translations of the original non-English questions.
The vast majority of our observed cases are related to weak cross-lingual encoding on the baseline model's part, where at least one important non-English word/entity in the question seems to be incorrectly matched with a similar but different English entity in the passage (e.g., the name of a different place).
For the Korean, Russian and Arabic queries in the table, we observe the presence of such topically similar entities (\eg{}, \textit{microwave} $\leftrightarrow$ \textit{gamma-ray}, \textit{Germany} $\leftrightarrow$ places in North America).
Much more rarely, we see cases similar to the Japanese query where the retrieved passage is completely off-topic.

\begin{table*}
\centering
\small
\begin{tabular}{p{14cm}}
\textcolor{darkblue}{\textbf{\textit{Korean Query}}}: 우주 마이크로파 배경은 누가 처음 발견했나요?\\
\textcolor{darkblue}{\textbf{\textit{English Translation}}}: Who first discovered the cosmic microwave background? \\
\textcolor{darkblue}{\textbf{\textit{Answer}}}: Arno Penzias and Robert Wilson \\
\textcolor{darkblue}{\textbf{\textit{Baseline Top Passage:}}} History of gamma-ray burst research The history of gamma-ray began with the serendipitous detection of a gamma-ray burst (GRB) on July 2, 1967, by the U.S. Vela satellites. After these satellites detected fifteen other GRBs, Ray Klebesadel of the Los Alamos National Laboratory published the first paper on the subject, "Observations of Gamma-Ray Bursts of Cosmic Origin". As more and more research was done on these mysterious events, hundreds of models were developed in an attempt to explain their origins. Gamma-ray bursts were discovered in the late 1960s by the U.S. Vela nuclear test detection satellites. The Velas were built \\
\textcolor{darkblue}{\textbf{\textit{\drdecr{} Top Passage:}}} project to date, it consists of more than 66 dishes and was built in an international collaboration by Europe, North America, East Asia and Chile. A major recent focus of microwave radio astronomy has been mapping the cosmic microwave background radiation (CMBR) discovered in 1964 by radio astronomers \hl{Arno Penzias and Robert Wilson}. This faint background radiation, which fills the universe and is almost the same in all directions, is "relic radiation" from the Big Bang, and is one of the few sources of information about conditions in the early universe. Due to the expansion and thus cooling of the \\
\hline
\textcolor{darkblue}{\textbf{\textit{Russian Query}}}: В каком сражении участвовал крейсер "Аврора"? \\
\textcolor{darkblue}{\textbf{\textit{English Translation}}}: What battle did the cruiser Aurora take part in? \\
\textcolor{darkblue}{\textbf{\textit{Answer}}}: Battle of Tsushima \\
\textcolor{darkblue}{\textbf{\textit{Baseline Top Passage:}}} Battle of the Arar The Battle of the Arar was fought between the migrating tribes of the Helvetii, and four Roman legions (Legions VII, VIII, IX "Hispana" and X "Equestris"), under the command of Gaius Julius Caesar, in 58 BC. This was the first major battle of the Gallic Wars. The Helvetii were a tribe that originated from what is now Switzerland. Just prior to the battle with Caesar, they had commenced on a mass migration through Roman Gaul towards the Atlantic coast. At Geneva, the Romans destroyed the wooden bridge across the Rhone and constructed 19 mi of fortifications. \\
\textcolor{darkblue}{\textbf{\textit{\drdecr{} Top Passage:}}} ordered back to Asia as part of the Russian 2nd Pacific Squadron, a collection formed from the Russian Baltic Fleet, under the command of Vice-Admiral Zinovy Rozhestvensky. On the way to the Far East, "Aurora" sustained light damage from confused friendly fire, which killed the ship's chaplain and a sailor, in the Dogger Bank incident. On 27 and 28 May 1905 "Aurora" took part in the \hl{Battle of Tsushima}, along with the rest of the Russian squadron. During the battle her captain, Captain 1st rank Eugene R. Yegoryev, and 14 crewmen were killed. The executive officer, Captain of 2nd rank \\
\hline
\textcolor{darkblue}{\textbf{\textit{Japanese Query}}}: 抑圧された記憶を提唱したのは誰 \\
\textcolor{darkblue}{\textbf{\textit{English Translation}}}: Who proposed repressed memory \\
\textcolor{darkblue}{\textbf{\textit{Answer}}}: Sigmund Freud \\
\textcolor{darkblue}{\textbf{\textit{Baseline Top Passage:}}} 14 on the charts on another reissue at the end of 1979, and is now regarded as the Moody Blues signature song by British audiences. In the US, "Nights in White Satin" did not make the "Billboard" Hot 100 in 1968, although it reached No. 2 on re-release in 1972; "Tuesday Afternoon" was more successful on initial release stateside, peaking at No. 24 on the Billboard Hot 100. The 1968 follow-up LP, "In Search of the Lost Chord" included "Legend of a Mind", a song written by Ray Thomas in tribute to LSD guru Timothy Leary which encompassed a flute \\
\textcolor{darkblue}{\textbf{\textit{\drdecr{} Top Passage:}}} literary example of traumatic memory that has been repressed by an otherwise healthy individual, and then recovered." Pope claimed that no entrant had satisfied the criteria. Ross Cheit, a political scientist at Brown University, cited Nina, a 1786 opera by the French composer Nicolas Dalayrac. The concept of repressed memory originated with \hl{Sigmund Freud} in his 1896 essay "Zur Ätiologie der Hysterie" ("On the etiology of hysteria"). One of the studies published in his essay involved a young woman by the name of Anna O. Among her many ailments, she suffered from stiff paralysis on the right side of her \\
\hline
\textcolor{darkblue}{\textbf{\textit{Arabic Query}}}: \<ما هو اكبر اقاليم المانيا؟>\\
\textcolor{darkblue}{\textbf{\textit{English Translation}}}: What is the largest region of Germany? \\
\textcolor{darkblue}{\textbf{\textit{Answer}}}: Bavaria \\
\textcolor{darkblue}{\textbf{\textit{Baseline Top Passage:}}} the original name of Montana was adopted. Montana is one of the nine Mountain States, located in the north of the region known as the Western United States. It borders North Dakota and South Dakota to the east. Wyoming is to the south, Idaho is to the west and southwest, and three Canadian provinces, British Columbia, Alberta, and Saskatchewan, are to the north. With an area of , Montana is slightly larger than Japan. It is the fourth largest state in the United States after Alaska, Texas, and California; it is the largest landlocked U.S. state. The state's topography is \\
\textcolor{darkblue}{\textbf{\textit{\drdecr{} Top Passage:}}} Bavaria (; German and Bavarian: "Bayern" ; ), officially the Free State of Bavaria (German and Bavarian: "Freistaat Bayern" ), is a landlocked federal state of Germany, occupying its southeastern corner. With an area of 70,550.19 square kilometres (27,200 sq mi), \hl{Bavaria} is the largest German state by land area. Its territory comprises roughly a fifth of the total land area of Germany. With 13 million inhabitants, it is Germany's second-most-populous state after North Rhine-Westphalia. Bavaria's capital and largest city, Munich, is the third-largest city in Germany. The history of Bavaria stretches from its earliest settlement and formation as \\
\hline
\caption{
Examples of cases where the baseline model fails to retrieve a relevant passage but \drdecr{} succeeds within top 3.
We only show the top retrieval for each system.
Most errors are related to potential word/entity mistranslations, the only exception being the Japanese query where the issue is a weaker understanding of the passage content.
}
\label{table:qualitative}
\end{tabular}
\end{table*}

\end{document}